\documentclass{sig-alternate-05-2015}

\begin{document}

% Copyright
% \setcopyright{acmcopyright}
%\setcopyright{acmlicensed}
%\setcopyright{rightsretained}
%\setcopyright{usgov}
%\setcopyright{usgovmixed}
%\setcopyright{cagov}
%\setcopyright{cagovmixed}

% DOI
% \doi{10.475/123_4}
% \doi{000000}

% ISBN
% \isbn{123-4567-24-567/08/06}
% \isbn{000000}

%Conference
% \conferenceinfo{WSDM '17}{February 10, 2017, Cambridge, UK}

% \acmPrice{\$15.00}

%
% --- Author Metadata here ---
% \conferenceinfo{WSDM}{2017 Cambridge, UK}
% \CopyrightYear{2007} % Allows default copyright year (20XX) to be over-ridden - IF NEED BE.
% \crdata{0-12345-67-8/90/01}  % Allows default copyright data (0-89791-88-6/97/05) to be over-ridden - IF NEED BE.
% --- End of Author Metadata ---

\title{Learning from various labeling strategies for suicide-related messages on social media: \\An experimental study}


\numberofauthors{5}
\author{
% 1st. author
\alignauthor
Tong Liu \\
      \affaddr{Golisano College of Computing and Information Sciences}\\
      \affaddr{Rochester Institute of Technology}\\
      \affaddr{Rochester, NY USA}\\
      \email{tl8313@rit.edu}
% 2nd. author
\alignauthor
Qijin Cheng \\
      \affaddr{HKJC Centre for Suicide Research and Prevention}\\
      \affaddr{The University of Hong Kong}\\
      \affaddr{Pok Fu Lam, Hong Kong}\\
      \email{chengqj@connect.hku.hk}
\and  % use '\and' if you need 'another row' of author names
% 3rd. author
\alignauthor 
Christopher M. Homan \\
      \affaddr{Golisano College of Computing and Information Sciences}\\
      \affaddr{Rochester Institute of Technology}\\
      \affaddr{Rochester, NY USA}\\
      \email{cmh@cs.rit.edu}
% 4th. author
\alignauthor Vincent M.B. Silenzio \\
      \affaddr{Department of Psychiatry}\\
      \affaddr{University of Rochester Medical Center}\\
      \affaddr{Rochester, NY USA}\\
      \email{vincent$\_$silenzio@urmc.\\rochester.edu}
}

\maketitle
\begin{abstract}
Suicide is an important but often misunderstood problem, one that researchers are now seeking to better understand through social media. Due in large part to the fuzzy nature of what constitutes suicidal risks, most supervised approaches for learning to automatically detect suicide-related activity in social media require a great deal of human labor to train. However, humans themselves have diverse or conflicting views on what constitutes suicidal thoughts. So how to obtain reliable gold standard labels is fundamentally challenging and, we hypothesize, depends largely on what is asked of the annotators and what slice of the data they label. We conducted multiple rounds of data labeling and collected annotations from crowdsourcing workers and domain experts. We aggregated the resulting labels in various ways to train a series of supervised models. Our preliminary evaluations show that using unanimously agreed labels from multiple annotators is helpful to achieve robust machine models.
\end{abstract}




%
%  Use this command to print the description
%

% We no longer use \terms command


\section{Introduction}

Social media provides a public lens into the daily lives and personal emotions of its users. Since people sometimes post about suicidal or self-harm-related thoughts, social media such as Facebook have established suicide reporting and prevention mechanisms via specific links and buttons to let users report when they or their friends encounter direct life threats. Though this is a big step in the right direction, it alone does little to advance our scientific understanding of suicide, or how to predict its likelihood. 

Researchers are increasingly using machine learning techniques to observe and study a wide range of mental health problems from social media activities. Most existing approaches rely heavily on data that are increasingly annotated by crowdsourced workers. Obtaining labels for issues this way for topics as subjective as mental health is challenging, as both crowdsourcing workers and domain experts often have different interpretations of the texts they read and label. Thus having multiple annotators look at and annotation each data item is absolutely.

When aggregating these multiple annotations into intelligent models for prevention purposes, how the training corpora is constructed---more specifically, what final ground truth labels researchers determine to apply into the supervised algorithms---will make significant differences to the final performance of output models. But to our best knowledge, there are few studies that have investigated how this kind of social media data were annotated and used differently in supervised learning settings. 

In this paper, we study approaches for annotation and label aggregation to train a classifier to detect suicide-related tweets gathered from Twitter using keyword- and location- based queries. Crowdsourced workers perform the first round of annotations, and then experts (the authors) annotated those tweets on which the crowdsourced workers disagreed. Our underlying assumption is that crowdsourced annotations with high inter-annotator agreement can provide us reliable usable corpora in supervised modeling of automatic classifiers. We experimented with different ways for aggregating multiple annotations into gold-standard labels and compared the effects of each. Our main contributions are:

\begin{itemize}
  \item We perform multiple rounds of annotations, involving both crowdsourcing workers and experts, to provide a diverse set of judgements about suicide-related discourse on social media data.
  \item We propose a variety of strategies for selecting the data to annotate and for determining ground truth labels in the succeeding training phase.
  \item We conduct experiments to compare---in terms of how each  method affects machine learning performance---sets of training data labeled by different method.
  \item We demonstrate that a labeling method based only on  unanimously labeled data from both crowdsourced workers and experts is the best way to train supervised models, especially when the problem domain is subjective.
\end{itemize}

\section{Related Work}

%In many natural language processing settings linguistic corpora are often built by linguists or crowdsourcers. Their labels are then simply taken as the gold standard labels and used in training and testing stages. However, in computational social science area, due to the the difficulty of determining ground truth labels in various problems, currently there are no standard methods for solving this issue.

%The wide spread of suicidal ideation in social media not only does harm to those people in life-threatening situations, but also poses potential risks to vulnerable people and subsequently come to harm to the community. This raised many researchers' awareness and concerns to build intelligent models to identify these worrying contents online. 

% CMH the above lines are not really related work
% Also, I'm told there's a paper by Callison-Burch on 
Kumar et al. \cite{kumar2015detecting} presented their research on the prevalence of Werther effect (copycat suicides following a celebrity's suicide) in social media. They examined posting activities and contents after public figures' suicide, and observed a virtual analog of sorts of the Werther effect: that people post more frequently with expressions of suicidal tendencies, and the linguistic measures change towards negative and biased direction.

``The Durkheim Project'' \cite{poulin2014predicting} further analyzed an opt-in database of veterans` social media and mobile phone data to seek real-time assessments and predictive analytics for psychological suicide risk factors. They developed text-based prediction models from single keywords and multi-word phrases, achieved about 65\% accuracy in identifying statistically significant signals of suicidality, and suggested the usefulness of computerized textual analytics of social media data to estimate the risk of suicide. But the group did not give details about the data -- how they determined the labels for training data to build the system. 

In \cite{burnap2015machine}, Burnap et al. focused on building a multi-class machine classifier with competitive accuracy when assigning tweets to particular class of suicidal communications and provided more details about their experiments. They requested at least four annotations per tweet from a crowdsourcing platform in order to limit the amount of subjectivity in the process of labeling suicidal tweets and kept tweets with high agreement (> 75\% -- at least three out of four annotators agreed on the dominant class of each tweet) to train classification models. Their empirical findings suggest that it is feasible for crowdsouring workers who are unknown to each other and without being influenced by each other`s judgement to reach agreement on the disclosure of suicidal ideation. Also based on the experiments and conclusions from \cite{snow2008cheap}, non-expert contributors could produce comparable quality of annotations when evaluating against those gold standard annotations from experts. And it is similarly effective to use the labeled tweets with high inter-annotator agreement among multiple non-expert annotators from crowdsourcing platforms to build robust models as doing so on expert-labeled data. 
% Our conjecture about crowdsourced annotations, based on the experiments and conclusions from \cite{snow2008cheap}, is that non-expert contributors could produce comparable quality of annotations when evaluating against those gold standard annotations from experts. And it is similarly effective to use the labeled tweets with high inter-annotator agreement among multiple non-expert annotators from crowdsourcing platforms to build robust models as doing so on expert-labeled data. 

\section{Data}

\subsection{Twitter Sampled Data -- Source 1}

Inspired by \cite{kumar2015detecting}, we searched for historical Twitter posts worldwide that were related to Robin Williams's suicide case and relevant information about suicide preventions using seven keywords and phrases suggested by suicide prevention experts and social workers, which are ``\textit{Robin Williams}'', ``\textit{suicide}'', ``\textit{depression}'', ``\textit{Parkinson's disease}'', ``\textit{seek help}'', ``\textit{suicide lifeline}'', and ``\textit{crisis hotline}''. We downloaded ten percent of Twitter messages that covered the scope of six months before and after Robin Williams' death (August 11th, 2014) and contained at least one of the above terms via DataSift API\footnote{http://datasift.com/}. This random sampling yielded approximately 1.7 million unique tweets in English from public accounts all over the world.

\subsection{Twitter Regional Data -- Source 2}
% For comparison purposes, 
We took, as a representative sample of typical Twitter use, historical Twitter data from three metropolitan centers in the United States
%(New York City, Rochester, Detroit) 
that  cover %both urban and rural 
a range of population densities. Most of the tweets in this set are in English.

\section{Annotations}

\subsection{Annotation Task Design}

We examined and borrowed a series of pattern matching rules from \cite{burnap2015machine} to generate many Twitter posts which are possibly related to suicide ideation or suicidal thoughts. This initial rule-based filter acts as our first classification model ($C_0$) that extracts suicide-related posts. $C_0$ searches for a wide range of expressions which include: \emph{suicidal / depression / cutting / bad / sad / these ... thoughts / feelings}, \emph{want / wanted / wanting to die}, \emph{end / ending it all}, \emph{end my life}, \emph{can't take (it) anymore}, \emph{can't / don't want to live any more}, \emph{don't want to be alive}, \emph{can't go on}, \emph{call / ask for help}, \emph{offer of help}, \emph{stop bullying}, \emph{kill / killing / hate myself}, \emph{fuck / fucking}, \emph{boyfriend / girlfriend}, \emph{just ... like}, \emph{talk / speak to someone / somebody}, \emph{web / blog / health  / advice}, \emph{miss / missing you / her / him}, \emph{took / taken (my / your / his / her) own life}, \emph{hanged / hanging / overdose}, etc. 

We ran $C_0$ on our pooled dataset and randomly selected 2,000 matched tweets (1,200 tweets from \emph{source 1} and 800 from \emph{source 2}) for manual annotations and validations. In particular, we anonymized the data to minimize the disclosure of personal information ($@$names) or URLs that may reveal cues about users' online identities.

% \begin{table}[htbp]
% \centering
% \begin{tabular}{c}
% suicidal/depression/cutting/bad/sad/these...thoughts/feelings \\ \hline
% want/wanted/wanting to die \\ \hline
% end/ending it all, end my life \\ \hline
% \begin{tabular}[c]{@{}c@{}}can't take (it) anymore, can't/don't want to live/cope \\ any more, don't want to be alive, can't go on\end{tabular} \\ \hline
% call/offer for/of help \\ \hline
% stop bullying \\ \hline
% kill/killing/hate myself \\ \hline
% fuck/ fucking \\ \hline
% boyfriend/girlfriend \\ \hline
% just...like \\ \hline
% talk/speak to someone/somebody \\ \hline
% web blog/health advice \\ \hline
% miss/missing you/her/him \\ \hline
% took/taken (my/your/his/her) own life \\ \hline
% hanged/hanging/overdose \\ 
% \end{tabular}
% \caption{Example phrases in $C_0$}
% \label{regex_terms}
% \end{table}

\subsection{Round1: Crowdsourced Annotations}

We first published this combination of 2,000 tweets on CrowdFlower\footnote{\url{https://www.crowdflower.com/}: This is an Amazon Mechanical Turk type crowdsourcing platform. Its software as a service platform allows requesters to access online workforce to clean, label and enrich data.}, five tweets per page, to invite workers to finish the labeling tasks as instructed. For each tweet, five annotators were paid fairly to choose only one label to best describe the category from four given choices (with one sentence in following parentheses to provide more descriptions):

\begin{itemize}
    \item \textbf{A.} Suicidal thoughts (The author or the author's friend is at risk of suicide/distress.)
    \item \textbf{B.} Supportive messages or helpful information (The author is providing supportive messages/helpful information related to suicide/distress.)
    \item \textbf{C.} Reaction to suicide news/movie/music (The author is spreading/reacting/commenting to suicide news/\\movie/music.)
    \item \textbf{D.} Other (The author is using suicide/distress words to describe something else.)
\end{itemize}

The rationale behind the design of multiple choice questions is: our data collection method (\emph{source 1} especially) inevitably introduced tweets covering topics such as category B or C among four choices, which are not necessarily our focus on the personal suicidal disclosures detection in this case study. At the same time, this setting is useful to manually reduce the complexities of automatic classification: Annotators intuitively differentiate the contents so that form some explicit boundaries between the target class (\emph{suicidal}) and noises before passing data into the supervised learning algorithms in the following classifier modeling phase.

\subsubsection{System Aggregated Labels ($R_1S$)}

CrowdFlower by default automatically aggregated five responses into a summarized result for each tweet based on the majority vote of the trusted workers.
%We obtained 173 tweets annotated as ``A. suicidal thoughts'', 265 tweets as ``B. supportive messages or helpful information'', 523 tweets as ``C. reaction to suicide news/movie/music'', and the rest 1039 as ``D. other''. 

\subsubsection{Unanimous Voted Labels ($R_1U$)}

There are 415 tweets labeled with unanimous agreement among five workers, i.e. five workers gave the same label to one tweet. The remaining 1,585 tweets were not labeled unanimously which have lower inter-annotator agreement.

\subsubsection{Observation}

The percentage of tweets with unanimous labels in this annotation round (20.75\%) is much smaller than that of some published experiments using the similar annotation strategy but for different social issues: Liu et al. asked crowdsourcing workers to differentiate job-themed tweets from the rest topics in \cite{liu2016understanding}, and harvested more than half of their published tweets with full agreement among five annotators per tweet (64.85\%). This observation and comparison intuitively suggests that non-expert crowdsourcing workers working on suicide-related annotation tasks have diverse types of understanding about this topic because of its sensitivity and ambiguity. We examined this part of data which were interpreted differently among five annotators per message in the subsequent section.

\subsection{Round2: Expert Annotations}

Experts were introduced into this phase to actively inspect what kinds of tweets that cause the divergent opinions from crowdsourcing workers. The 1,585 tweets with diversified labels from Round1 were then published internally to two experts to have them labeled twice. The tweets with unanimous labels from crowdsourcing workers were not re-annotated by experts because unanimous votes are hypothesized to be reliable as expert' labels.

\subsubsection{Determining the Labels}

We assigned the identical labels from the two experts to the gold standard label for each tweet ($R_2U$). We compromised for those tweets annotated differently by two experts by adopting the system aggregated labels from Round1 crowdsourced annotations ($R_2S$).

\subsection{Annotations Summary}

Table \ref{annotation_records} records the percentages of tweets in four categories (A, B, C, and D) collected in different sources of annotations, with five distinct strategies to determine the final ground truth label for each tweet.

\begin{table}[htbp]
\centering
\caption{Statistics of labels obtained from different sources of annotations. \textbf{$R_1$}: Crowdsourced annotations. \textbf{$R_2$}: Expert annotations. \textbf{S}: Each tweet label comes from the system aggregation following the majority vote rules. \textbf{U}: Each tweet label is the unanimously voted choice among five annotators. \textbf{+}: Union operation to combine elements in each annotation set. \textbf{Total}: The actual counts of tweets.}
\label{annotation_records}
\begin{tabular}{c|c|c|c|c|c}
\textbf{Category (\textbf{\%})} & \textbf{A.} & \textbf{B.} & \textbf{C.} & \textbf{D.} & \textbf{Total} \\ \hline
$R_1S$ & 8.65 & 13.25 & 26.15 & 51.95 & \textbf{2,000} \\ 
%\hline
$R_1U$ & 2.89 & 8.67 & 17.11 & 71.33 & \textbf{415} \\ 
%\hline
$R_2U$ & 5.37 & 19.48 & 31.29 & 43.86 & \textbf{1,042} \\ 
%\hline
$R_1U$ + $R_2U$ & 4.67 & 16.40 & 27.25 & 51.68 & \textbf{1,457} \\ 
%\hline
$R_1U$ + $R_2U$ + $R_2S$ & 7.85 & 16.00 & 26.55 & 49.60 & \textbf{2,000} \\ 
%\hline
\end{tabular}
\end{table}

% \begin{table}[htbp]
% \centering
% \begin{tabular}{c|c|c|c|c||c}
% \textbf{Category} & \textbf{A.} & \textbf{B.} & \textbf{C.} & \textbf{D.} & \textbf{Total} \\ \hline
% \hline
% $R_1S$ & 173 & 265 & 523 & 1,039 & 2,000 \\ \hline
% $R_1U$ & 12 & 36 & 71 & 296 & 415 \\ \hline
% $R_2U$ & 56 & 203 & 326 & 457 & 1,042 \\ \hline
% $R_1U$ + $R_2U$ & 68 & 239 & 397 & 753 & 1,457 \\ \hline
% $R_1U$ + $R_2U$ + $R_2S$ & 157 & 320 & 531 & 992 & 2,000 \\ \hline
% \end{tabular}
% \end{table}

\begin{figure}[h]
\centering
\includegraphics[scale=0.5]{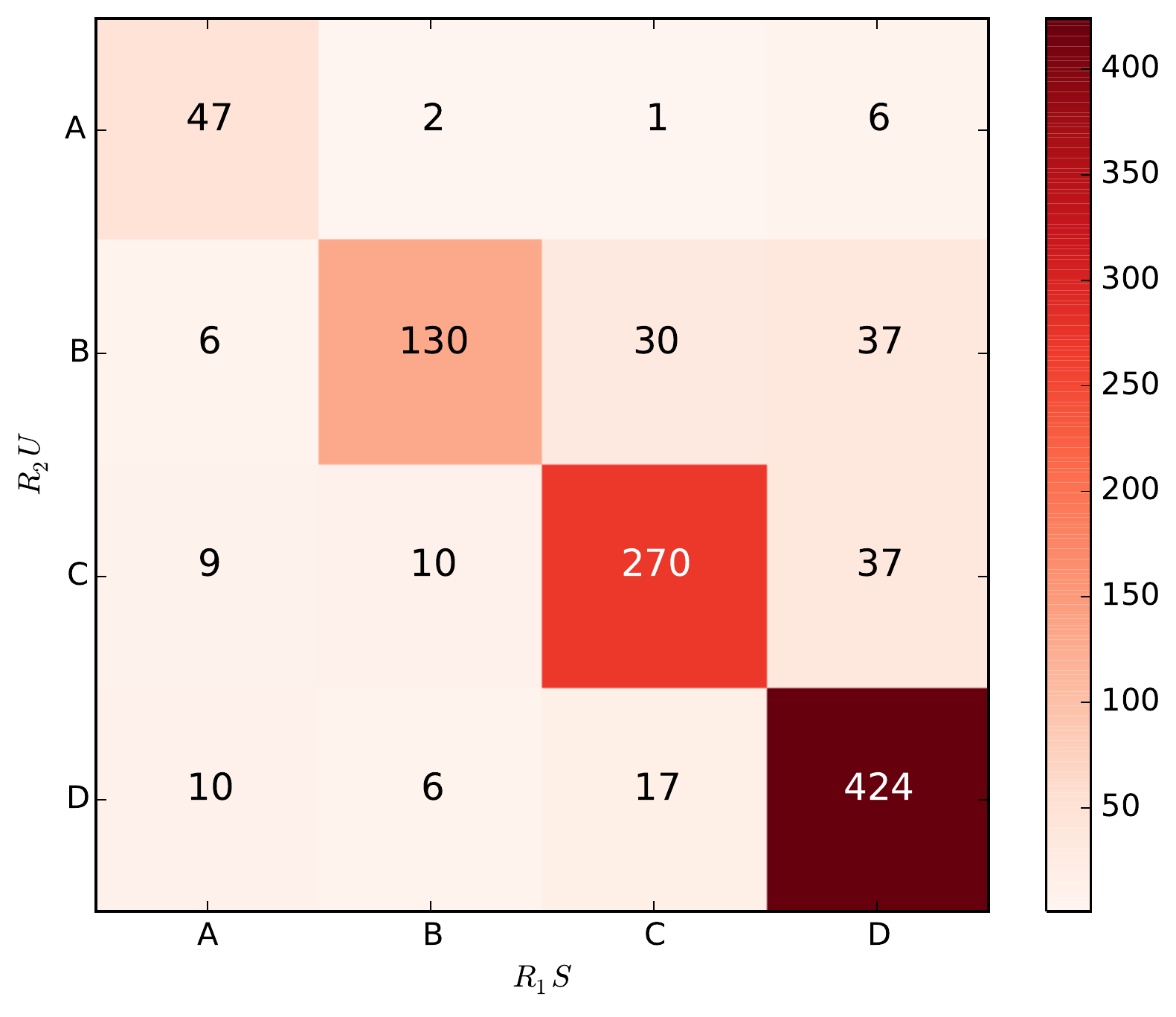}
\caption{Comparisons between $R_1S$ and $R_2U$.}
\label{R1SR2U}
\end{figure}

For the tweets with gold standard unanimous labels from experts ($R_2U$ in Table \ref{annotation_records}), we compared them to the labels aggregated from the crowdsourced annotations ($R_1S$) and found a total of 871 common tweets having the same annotation results, which is approximately 83.59\% of the number of tweets with unanimous agreement between two experts. Among them, 47 tweets belong to ``A. suicidal thoughts'', 130 ``B. supportive messages or helpful information'', 270 ``C. reaction to suicide news/movie/music'' and 424 ``D. other'' (see the comparisons between $R_1S$ and $R_2U$ in Figure \ref{R1SR2U}).

\begin{figure}[h]
\centering
\includegraphics[scale=0.5]{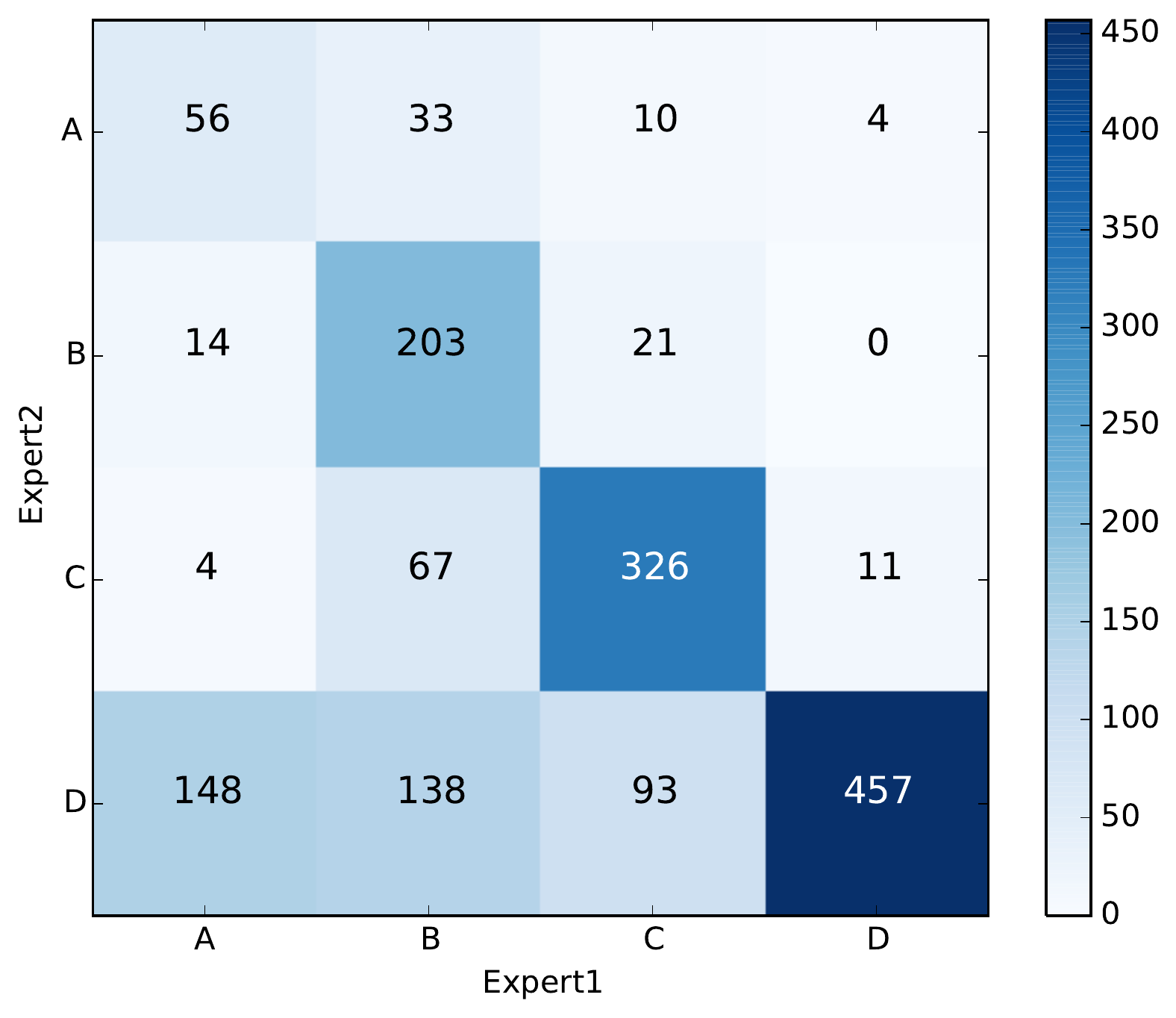}
\caption{Comparisons between two expert annotators in Round2.}
\label{R2}
\end{figure}

In Figure \ref{R2} we further compared the annotations between two experts in Round2. We assessed their inter-annotator agreement on multiple choice labeling tasks using Cohen's kappa \cite{cohen1960coefficient} as $\kappa = 0.523$. 

In Table \ref{sample_tweets}, we presented several samples from two rounds of annotations with their labels contributed by crowdsourcing workers ($R_1$) and experts ($R_2$). 

\begin{table*}[]
\centering
\caption{Sample tweets with annotations from crowdsourcing workers and experts in two rounds.}
\label{sample_tweets}
\begin{tabular}{c|c|c}
% \hline
\textbf{$R_1$ annotations} & \textbf{$R_2$ annotations} & \textbf{Sample message} \\ \hline
AAAAA & --- & \textit{\begin{tabular}[c]{@{}c@{}}Wonder if I died and am in hell...depression is eating at me\\  bad last couple of days...\end{tabular}} \\ \hline
BBBBB & --- & \textit{@SOMEONE wishing you good health and happiness} \\ \hline
AAAAC & AD & \textit{\begin{tabular}[c]{@{}c@{}}To me, suicide seems selfish. For all I know, someone else might\\  want to kill me\end{tabular}} \\ \hline
AADDD & AD & \textit{\begin{tabular}[c]{@{}c@{}}Something happens every time. I honestly don't know why I even \\ bother anymore. Fuck this. \#useless\end{tabular}} \\ \hline
ABCCD & CC & \textit{\begin{tabular}[c]{@{}c@{}}Having someone I considered my brother commit suicide last year had to be\\  one of the most eye-opening experiences in my life.\end{tabular}} \\ \hline
AAADD & AA & \textit{I'd rather kill myself than commit suicide} \\ \hline
BCCCD & CC & \textit{\begin{tabular}[c]{@{}c@{}}A boy from my school committed suicide today and the boy who committed\\  over the summer's birthday is today\end{tabular}} \\ \hline
AAADD & AD & \textit{\begin{tabular}[c]{@{}c@{}}I really hate being a girl. In 3 hours today I went from excited to pissed to sad \\ to happy. Don't look at me wrong or I might cry. \#dafaq?\end{tabular}} \\ \hline
CCCCD & BC & \textit{\begin{tabular}[c]{@{}c@{}}According to a British law passed in 1845, attempting to commit suicide\\  was a capital offence. Offenders could be hanged for trying.\end{tabular}} \\ 
%\hline
\end{tabular}
\end{table*}

This table exemplifies the complexity of labeling suicide-related tweets among multiple annotators. For example, tweets like ``\emph{According to a British law passed in 1845, attempting to commit suicide was a capital offence. Offenders could be hanged for trying.}'' declares that suicide was criminalized in the past. It might reinforce the stigma of suicide and make people unwilling to seek help timely, or can be understood as some way to raise the public awareness of suicide. Giving another example, some tweets are talking about what depression feels like without any explicit phrases about depression or suicide prevention. Some annotators classified such messages as ``Supportive messages or helpful information'' while they may actually worsen another person's existing negative feelings. Tweets mentioning suicide bombing/attack is an additional category about which annotator have diverse understandings. Such ambiguities make annotators, even experts, label such kind of tweets differently.

\section{Modeling Experiments}

To simplify the modeling process to identify tweets which express personal suicide ideation and suicidal thoughts and differentiate between this and other types of suicide-related messages, we grouped tweets with labels in category B, C and D into one class and formed the data points into binary categories: \emph{suicidal} (positive) vs. \emph{others} (negative).

Five combinations of data in Table \ref{annotation_records} were entered in our feature extraction and modeling pipeline to study the influence of different labeling strategies to classification modeling performances.

\subsection{Model}

To control the environment variables of this experimental study, we selected support vector machines (SVMs) as our supervised learning methods to build a series of classification models. An SVM model takes in a set of training data, each labeled as belonging to one specific category, forms an optimal separating hyperplane to maximize the margin of input training data that are represented as data points in feature space. This algorithm outputs a discriminative classifier which is able to categorize new examples (provide a predicted class label) after mapping them into the same feature space. We used the scikit-learn implementation \cite{pedregosa2011scikit} of SVMs in the experiments.

\subsection{Feature Preparation}

We relied on the textual representations (N-grams) to train and evaluate a series of SVM classifiers. Due to the noisy nature of Twitter, where people frequently write short, informal spellings and grammars, we pre-processed tweets as the following steps: (1) replaced personal information ($@$names) with \emph{$@SOMEONE$}, and recognizable URLs with \emph{$HTTP://LINK$}, (2) utilized a revised \texttt{Twokenizer} system which was specially trained on Twitter texts \cite{owoputi2013improved} to tokenize raw messages, and (3) completed stemming and lemmatization using WordNet Lemmatizer \cite{bird2009natural}.

The statistics of N-grams (unigrams, bigrams and trigrams) extracted from different sets of training data with mixed labeling strategies are summarized in Table \ref{features}. We used the top \textbf{10,000} unique N-grams as features in the modeling process of $C_1$ to $C_5$. 

\begin{table*}[ht]
\centering
\caption{Statistics of features from different sources of annotations, to train models $C_1$ to $C_5$.}
\label{features}
\begin{tabular}{cc|ccc|c}
\textbf{Input Data} & \textbf{Output Model} & \textbf{Unigrams (\%)} & \textbf{Bigrams (\%)} & \textbf{Trigrams (\%)} & \textbf{Total N-grams Count} \\ \hline
$R_1S$ & $C_1$ & 10.48 & 39.55 & 49.98 & 45,582 \\ 
%\hline
$R_1U$ & $C_2$ & 15.89 & 40.75 & 43.36 & 10,493 \\ 
%\hline
$R_2U$ & $C_3$ & 12.37 & 40.14 & 47.49 & 25,620 \\ 
%\hline
$R_1U$ + $R_2U$ & $C_4$ & 11.55 & 40.01 & 48.44 & 33,678 \\ 
%\hline
$R_1U$ + $R_2U$ + $R_2S$ & $C_5$ & 10.48 & 39.55 & 49.98 & 45,582 \\ 
%\hline
\end{tabular}
\end{table*}

% \begin{table*}[ht]
% \centering
% % \caption{Statistics of features from different sources of annotations, to train models $C_1$ to $C_5$.}
% % \label{features}
% \begin{tabular}{c|c||c|c|c||c}
% \textbf{Input Data} & \textbf{Output Model} & \textbf{Unigrams} & \textbf{Bigrams} & \textbf{Trigrams} & \textbf{Total N-grams} \\ \hline
% $R_1S$ & $C_1$ & 4,776 & 18,026 & 22,780 & 45,582 \\ % \hline
% $R_1U$ & $C_2$ & 1,667 & 4,276 & 4,550 & 10,493 \\ 
% % \hline
% $R_2U$ & $C_3$ & 3,168 & 10,285 & 12,167 & 25,620 \\ % \hline
% $R_1U$ + $R_2U$ & $C_4$ & 3,890 & 13,475 & 16,313 & 33,678 \\ 
% % \hline
% $R_1U$ + $R_2U$ + $R_2S$ & $C_5$ & 4,776 & 18,026 & 22,780 & 45,582 \\ 
% % \hline
% \end{tabular}
% \end{table*}

\subsection{Parameter Selection}

Considering the class imbalance situations in each training dataset, we determined the optimal learning parameters by grid-searching on a range of class weights for the positive and negative classes, and then chose the set which optimized the area under the receiver operating characteristic curve (AUC)\footnote{We tested a variety of objective tuning functions during the grid-search process and concluded that \emph{AUC} could achieve the best precision, recall and F1-score on the positive class.}. 

\subsection{Performance Evaluation}

\subsubsection{K-fold Cross-Validation}

\textit{K-fold cross-validation} relieves our straits that we do not have too many labeled data in total: if further partitioning the available data into three sets---training set, validation set and test set---would drastically reduce the number of data used for learning purpose, and that the testing results could possibly depend on a particular random split for the pair of training-validation sets due to the potential over-fitting risks: the parameters can be tweaked until the estimator achieves the optimal performance on the validation set which ultimately causes that our final evaluation results could no longer reflect the model's generalizability on the unseen data.
%In the basic k-fold cross-validation approach, the training set is split into k smaller subsets (in our experiments, k is set to \textbf{10}). For each of the k ``folds'', a model is first trained using \emph{k-1} of the folds as training data; this learned model is then validated on the remaining fold of the data (i.e., it is used as a test set to compute a performance measure). The final performance measure reported by k-fold cross-validation is the average of the performance values computed in the above k loops. 
Though this evaluation approach is potentially computationally expensive, it does not waste too much data which is a major advantage in solving problem where the number of labeled samples is very small like our case. In our experiments, K is set to \textbf{10}.

\subsubsection{Learning Curve}

A learning curve illustration shows the training scores and cross-validation scores of an estimator for varying numbers of training samples which helps us understand how much the benefits we could get by adding more training data. It is also a tool to understand whether the estimator suffers more from a bias error or a variance error during the modeling process\footnote{For an estimator, the bias error is its average error for different training sets. The variance reflects its sensitivity to varying numbers of training data.}.

\section{Results and discussions}

Having the core supervised learning algorithm and dimensions of features as constant variables, we conducted our experiments by training five SVM classifiers ($C_1$ to $C_5$) using five sets of training data described in Table \ref{features}. We analyzed their similarities and differences in the following subsections:

\subsection{Learning Curve}

\begin{figure*}[ht]
\centering
\includegraphics[scale=0.70]{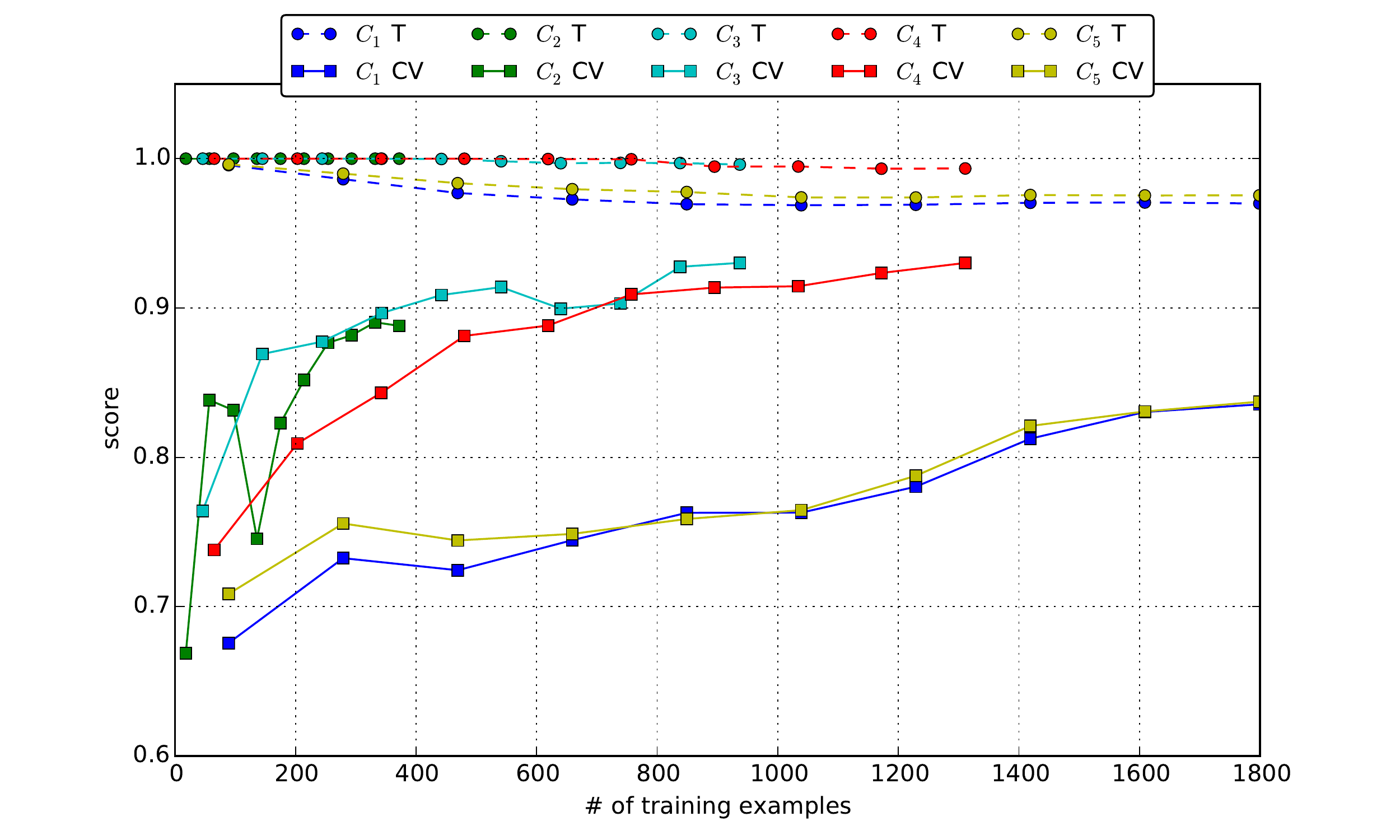}
\caption{Learning curves for models $C_1$ to $C_5$ during the training process. Dashed lines with circle markers represent training scores for each model, abbreviated as \textbf{T} in legend box. Solid lines with square markers represent cross-validation scores, noted as \textbf{CV}. $C_1$: blue; $C_2$: green; $C_3$: cyan; $C_4$: red; and $C_5$: yellow.}
\label{learning_curves}
\end{figure*}

Figure \ref{learning_curves} shows the learning curves for models $C_1$ to $C_5$ during the training process with training data gradually added.

% \begin{figure*}[h]
% \centering
% \includegraphics[width=.32\textwidth]{figures/C1_learning_curve_roc_auc.pdf}
% \includegraphics[width=.32\textwidth]{figures/C2_learning_curve_roc_auc.pdf}
% \includegraphics[width=.32\textwidth]{figures/C3_learning_curve_roc_auc.pdf}
% \medskip
% \includegraphics[width=.32\textwidth]{figures/C4_learning_curve_roc_auc.pdf}
% \includegraphics[width=.32\textwidth]{figures/C5_learning_curve_roc_auc.pdf}
% \caption{Learning curves for models $C_1$ to $C_5$ during the training process.}
% \label{learning_curves}
% \end{figure*}

We observed several similarities from Figure \ref{learning_curves}: (1) For each model, the trending line of training scores (dashed lines with circles) and that of the cross-validation scores (solid lines with squares) did not converge to a value that is too low with increasing size of the training set; and (2) The differences between training scores and cross-validation scores for each model are continuously great with more training samples added gradually as the X axis marked. Even at the point of the maximum number of training samples used, the training score is greater than the cross-validation score. These observations suggest that we will benefit from adding more training samples to increase its generalization performance as well as reduce its bias error for each model.

We also noticed in Figure \ref{learning_curves} that the cross-validation scores for five models reached different points on the Y axis when the maximum number of training data used, representing that they achieved different performances in the end. Among them, $C_4$ reaches the highest cross-validation score. $C_4$ also has the least variance errors according to our experiments, suggesting it has good stability over others. The cross-validation scores in $C_2$ increase and converge to the training score very quickly using the least small number of training samples among five models, but the fluctuations of cross-validation scores during the $C_2$ training process are significant, showing that its performance is not comparably stable. 

\subsection{Outstanding Features}

In Table \ref{C1_top_features} -- \ref{C5_top_features}, we present the top 20 features for both positive and negative classes of $C_1$ to $C_5$ respectively according to their weights in creating the separating hyperplane to break the suicidal tweets away from the opposite class based on the labels from different determination strategies. 

Comparing these features from $C_1$ to $C_5$, we observed that \emph{suicide}, \emph{depression} and \emph{feel} appear in the suicidal class of each model, suggesting that suicidal ideation is closely related to these signal words. \emph{Myself}, \emph{me}, \emph{kill myself}, \emph{my}, \emph{im}, \emph{my depression} show that suspected suicidal users focus more on themselves with frequent use of first person singular pronouns. This series of self-oriented linguistic characters match the definition of suicide that it is an act of intentionally taking one's own life and cause one's own death, and are consistent with previous research findings \cite{kumar2015detecting,rude2004language}. The primary root-word \emph{commit} shows up in both classes of each model, but in distinct verb tenses: people in suicidal class write \emph{committing suicide}, \emph{committing} frequently which are in present tense and describing their ongoing actions, while \emph{commits}, \emph{committed}, \emph{commits suicide} are used as common vocabularies (simple present tense) or past suicide tragedies (past tense). \emph{Thought} and \emph{tried (to)} are another set of commonly used words for users in suicidal class which implies that people expressed their thoughts or even attempts to end their own lives. \emph{Bomber}, \emph{suicide bomber}, \emph{robin williams}, and \emph{williams} appear as top features for non-suicidal tweets across a few datasets due to the fact that many tweets containing those phrases are reposting or commenting to suicide news.

\begin{table}[h!]
% \small
\centering
\caption{Top 20 features for both classes of $C_1$.}
\label{C1_top_features}
\begin{tabular}{c c|c c}
\textbf{Suicidal} & \textbf{weights} & \textbf{Others} & \textbf{weights} \\ \hline
suicide & 0.652 & your & -0.159 \\ 
myself & 0.425 & do & -0.155 \\ 
depression & 0.424 & bomber & -0.152 \\ 
die & 0.328 & suicide bomber & -0.137 \\ 
commit & 0.293 & if you & -0.131 \\ 
me & 0.282 & commits & -0.129 \\ 
suicide is & 0.270 & commits suicide & -0.129 \\ 
feel & 0.258 & williams & -0.128 \\ 
thought & 0.238 & suicide every & -0.119 \\ 
kill myself & 0.237 & health & -0.118 \\ 
not & 0.229 & of & -0.118 \\ 
my & 0.228 & them & -0.118 \\ 
commit suicide & 0.223 & after & -0.115 \\ 
committing suicide & 0.209 & isn & -0.108 \\ 
depression and & 0.208 & stop & -0.107 \\
fucking & 0.203 & committed & -0.106 \\
committing & 0.199 & advice & -0.106 \\
im & 0.196 & robin williams & -0.104 \\
an & 0.193 & his & -0.102 \\
suicidal & 0.190 & who & -0.102
\end{tabular}
\end{table}

\begin{table}[h!]
% \small
\centering
\caption{Top 20 features for both classes of $C_2$.}
\label{C2_top_features}
\begin{tabular}{c c|c c}
\textbf{Suicidal} & \textbf{weights} & \textbf{Others} & \textbf{weights} \\ \hline
suicide & 0.141 & someone & -0.063 \\ 
don & 0.121 & fuck & -0.038 \\ 
don wanna & 0.120 & on & -0.037 \\ 
this & 0.108 & with & -0.033 \\ 
also & 0.107 & who & -0.030 \\ 
also don & 0.107 & not & -0.030 \\ 
depression is eating & 0.107 & if you & -0.028 \\ 
is eating & 0.107 & love & -0.027 \\ 
eating & 0.106 & how & -0.027 \\ 
wanna & 0.099 & after & -0.026 \\ 
depression & 0.094 & your & -0.025 \\ 
depression is & 0.093 & chicago & -0.025 \\ 
commit & 0.092 & don like & -0.024 \\ 
commit suicide & 0.092 & don like chicago & -0.024 \\ 
tried & 0.090 & like chicago & -0.024 \\
anymore & 0.088 & get & -0.024 \\
much just & 0.087 & guy & -0.023 \\
am & 0.086 & need & -0.023 \\
day & 0.079 & out & -0.022 \\
feel like & 0.078 & committed & -0.022
\end{tabular}
\end{table}

\begin{table}[h!]
% \small
\centering
\caption{Top 20 features for both classes of $C_3$.}
\label{C3_top_features}
\begin{tabular}{c c|c c}
\textbf{Suicidal} & \textbf{weights} & \textbf{Others} & \textbf{weights} \\ \hline
myself & 0.422 & people & -0.119 \\ 
suicide & 0.385 & by & -0.113 \\ 
me & 0.342 & with & -0.103 \\ 
kill & 0.341 & be & -0.100 \\ 
kill myself & 0.292 & one & -0.085 \\ 
my & 0.230 & after & -0.083 \\ 
day & 0.228 & please & -0.078 \\ 
depression & 0.200 & you & -0.076 \\ 
feel & 0.168 & we & -0.074 \\ 
so & 0.156 & their & -0.067 \\ 
my depression & 0.153 & of & -0.067 \\ 
alive & 0.152 & year & -0.066 \\ 
rather & 0.150 & girl & -0.065 \\ 
im & 0.148 & need & -0.061 \\ 
commit & 0.148 & williams & -0.061 \\
tried to & 0.146 & committed & -0.060 \\
almost & 0.136 & today & -0.058 \\
tried & 0.130 & always & -0.058 \\
suicide is & 0.129 & he & -0.058 \\
cutting & 0.128 & get & -0.057
\end{tabular}
\end{table}

\begin{table}[h!]
% \small
\centering
\caption{Top 20 features for both classes of $C_4$.}
\label{C4_top_features}
\begin{tabular}{c c|c c}
\textbf{Suicidal} & \textbf{weights} & \textbf{Others} & \textbf{weights} \\ \hline
myself & 0.440 & with & -0.149 \\ 
suicide & 0.436 & people & -0.139 \\ 
me & 0.339 & by & -0.111 \\ 
kill & 0.314 & out & -0.104 \\ 
kill myself & 0.300 & after & -0.098 \\ 
depression & 0.251 & you & -0.083 \\ 
day & 0.233 & get & -0.077 \\ 
tried & 0.221 & we & -0.073 \\ 
im & 0.212 & need & -0.072 \\ 
alive & 0.194 & your & -0.071 \\ 
tried to & 0.191 & he & -0.070 \\ 
feel & 0.187 & be & -0.068 \\ 
my & 0.185 & girl & -0.068 \\ 
commit & 0.185 & boyfriend & -0.068 \\
last & 0.168 & bitch & -0.066 \\
so & 0.167 & always & -0.065 \\
rather & 0.165 & soldier & -0.065 \\
my depression & 0.161 & are & -0.061 \\
again & 0.156 & need to & -0.061 \\
everything & 0.148 & williams & -0.060
\end{tabular}
\end{table}

\begin{table}[h!]
% \small
\centering
\caption{Top 20 features for both classes of $C_5$.}
\label{C5_top_features}
\begin{tabular}{c c|c c}
\textbf{Suicidal} & \textbf{weights} & \textbf{Others} & \textbf{weights} \\ \hline
depression & 0.700 & of & -0.250 \\ 
suicide & 0.654 & health & -0.219 \\ 
myself & 0.418 & bomber & -0.139 \\ 
thought & 0.397 & after & -0.193 \\ 
feel & 0.386 & people & -0.186 \\ 
suicidal & 0.338 & williams & -0.168 \\ 
commit & 0.334 & suicide bomber & -0.167 \\ 
me & 0.294 & commits & -0.167 \\ 
die & 0.292 & commits suicide & -0.167 \\ 
suicide is & 0.278 & please & -0.160 \\ 
kill & 0.273 & self & -0.154 \\ 
depression is & 0.259 & photo & -0.151 \\ 
committing suicide & 0.258 & love & -0.148 \\ 
depression and & 0.255 & their & -0.146 \\ 
commit suicide & 0.254 & committed & -0.142 \\
committing & 0.248 & your & -0.142 \\
im & 0.237 & of depression & -0.140 \\
kill myself & 0.229 & robin williams & -0.138 \\
stress & 0.212 & advice & -0.133 \\
cry & 0.210 & on & -0.131
\end{tabular}
\end{table}

% \begin{table}[h!]
% % \small
% \centering
% \begin{tabular}{c c||c c}
% \textbf{Suicidal} & \textbf{weights} & \textbf{Others} & \textbf{weights} \\ \hline
% suicide & 0.524 & bomber & -0.143 \\ 
% myself & 0.459 & health & -0.140 \\ 
% depression & 0.456 & their & -0.139 \\ 
% commit & 0.287 & of & -0.138 \\ 
% me & 0.287 & after & -0.135 \\ 
% kill & 0.274 & people & -0.134 \\ 
% kill myself & 0.263 & suicide bomber & -0.131 \\ 
% commit suicide & 0.257 & who & -0.130 \\ 
% feel & 0.253 & williams & -0.125 \\ 
% suicide is & 0.245 & if you & -0.124 \\ 
% im & 0.239 & self & -0.121 \\ 
% thought & 0.239 & them & -0.115 \\ 
% my & 0.219 & of depression & -0.115 \\ 
% depression and & 0.207 & your & -0.109 \\ 
% now & 0.201 & please & -0.108 \\
% day & 0.201 & do & -0.103 \\
% tried & 0.193 & love & -0.101 \\
% suicidal & 0.190 & commits & -0.101 \\
% commiting suicide & 0.190 & commits suicide & -0.101 \\
% depression is & 0.181 & robin williams & -0.100
% \end{tabular}
% \caption{Top 20 features for both classes of $C_5$.}
% \label{C5_top_features}
% \end{table}

\subsection{Performance Evaluations}

\begin{figure}[h]
\centering
\includegraphics[scale=0.4]{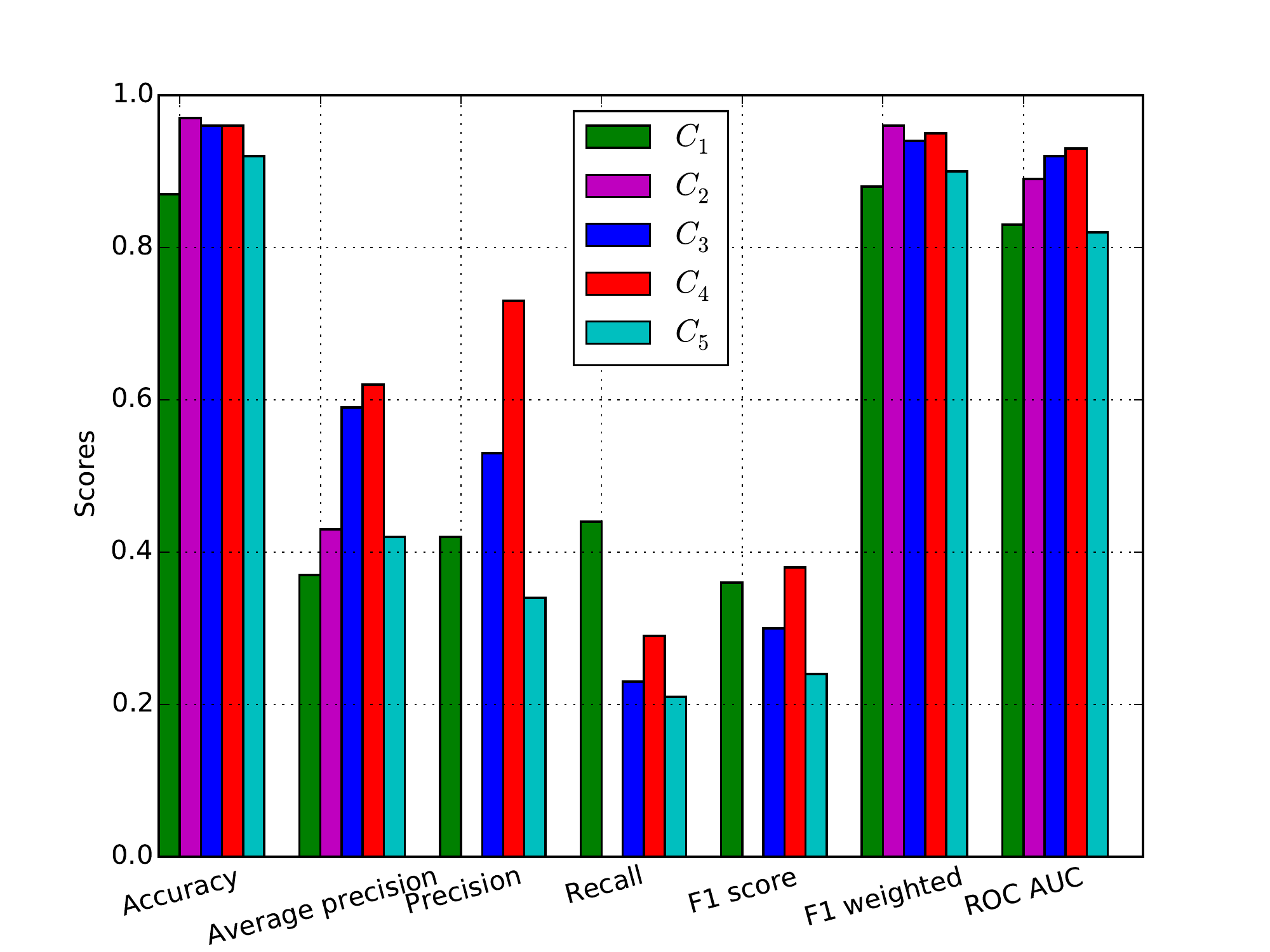}
\caption{Comparisons of performance metrics for $C_1$ to $C_5$.}
\label{metrics}
\end{figure}

Figure \ref{metrics} compares the five models according to seven performance metrics. We observed that $C_4$ stands out in \emph{average precision}\footnote{This score corresponds to the area under the precision-recall curve.}, \emph{precision}, \emph{F1 score} and \emph{ROC AUC} comparisons. The performances of $C_3$ are slightly lower than those of $C_4$ in most measures. $C_2$ has score 0 in \emph{precision}, \emph{recall} and \emph{F1 score} due to its bad performance which the number of correctly classified positives is 0 -- This result from $C_2$ is trained with the least size of training data, with only 12 positive samples. $C_2$ still has comparable performances in some measures and even surpasses scores of $C_4$ in \emph{accuracy} and \emph{F1 weighted score}\footnote{This measure accounts for class imbalance issue. It calculates metrics for each class and finds their average, weighted by the number of true instances for each class.}. This results from the greater disparity between positive and negative class in $R_1U$ than that in other training data. $C_1$ and $C_5$ generally achieved lower performance scores than other three models which were trained using only the unanimously labeled results from annotators (crowdsourcing workers, experts or their combination), suggesting some anti-correlations between lower inter-annotator agreement labels ($R_1S$ and $R_2S$) and the robustness of output models.

\section{Conclusions}

We raised a very foundational research question about determining the ground truth label for social media data before proceeding to building supervised classifiers, especially when the topic is sensitive, subjective and ambiguous. We controlled the settings of experiments (supervised learning algorithm and dimensions of features), and altered the input training data obtained from different labeling sources (crowdsourcing workers and experts) and strategies (majority votes and unanimous votes). We presented preliminary findings on the outcome of learning from multiple annotators on suicide-related annotation tasks. Our results show that it is helpful to use unanimous labels from crowdsourcing workers and experts as training data to build models. Though domain knowledge and experience are necessary in labeling the suicide-related data, our results provide some evidence that multiple crowdsourcing workers, when they reach high inter-annotator agreement, can provide reliable quality of annotations.

There are several interesting directions to future work. We did not investigate that how often the multiple workers unanimously agreed on the wrong outcome though it would be very uncommon. Examination of the annotated data without unanimous labels among multiple crowdsourcing workers will help understand where the disagreement is and propose solutions to improve public awareness and understanding of suicide so that we could rely more on crowdsourcing platforms to reduce the overall annotation costs. We also plan to investigate to what extent we could reduce the reliability loss of the disagreement among multiple annotators on the final output supervised models. Finally, given our observation of common heavily-weighted features expressed in suicidal posts, we might develop more complicated language models to automatically detect suicide ideation which could be helpful to provide decision making support to psychologists and psychiatrists, and ultimately care and support those vulnerable communities.

%\end{document}  % This is where a 'short' article might terminate

%ACKNOWLEDGMENTS are optional
\section{Acknowledgments}

We thank reviewers for their comments and suggestions. We are grateful to Cecilia Ovesdotter Alm, Rui Li, Ray Ptucha, Emily Prud'hommeaux and Xuan Guo for helpful conversations. This work was supported by US National Institutes of Health grant R25TW010012-01 and Hong Kong General Research Fund GRF17628916.

\bibliographystyle{abbrv}
\bibliography{sig-alternate-sample}  % sigproc.bib is the name of the Bibliography in this case
% ACM needs 'a single self-contained file'!
%

\balancecolumns % GM June 2007
% That's all folks!
\end{document}